\documentclass[letterpaper, 10 pt, conference]{ieeeconf}  
\IEEEoverridecommandlockouts                        
\usepackage{graphics} 
\usepackage{epsfig} 
\usepackage{mathptmx} 
\usepackage{times} 
\usepackage{amsmath} 
\usepackage{amsfonts} 
\usepackage{amssymb}  
\usepackage{float}
\usepackage{dblfloatfix}
\usepackage{bm}
\usepackage{tabularx}
\usepackage{booktabs}
\usepackage{tikzpagenodes}
\usepackage[hidelinks]{hyperref}
\usepackage{cite}

\usepackage{microtype}
\makeatletter
\MT@addto@setup{%
	\DeclareRobustCommand\microtypecontext[1]{%
		\MT@setup@contexts
		\let\MT@reset@context\relax
		\let\glb@currsize\@empty 
		\setkeys{MTC}{#1}%
		\selectfont
		\MT@reset@context
	}%
}
\makeatother

\usepackage[moderate]{savetrees}

\title{\LARGE \bf Online Adaptation for Myographic Control of\\Natural Dexterous Hand and Finger Movements} 

\author{Joseph L. Betthauser, Rebecca Greene, Ananya Dhawan, John T. Krall, Christopher L. Hunt,\\ Gy\"{o}rgy L\'{e}vay, Rahul R. Kaliki, Matthew S. Fifer, Siddhartha Sikdar, and Nitish V. Thakor
	\thanks{This work was supported by the Johns Hopkins University Applied Physics Laboratory Graduate Research Fellowship}
	\thanks{$^{}$J. Betthauser, R. Greene, and N. Thakor are with the Department of Electrical and Computer Engineering, The Johns Hopkins University, Baltimore, MD 21218, USA, {\small jbettha1@alumni.jh.edu}}%
	\thanks{$^{}$J. Krall, C. Hunt, and N. Thakor are with the Department of Biomedical Engineering, The Johns Hopkins University, Baltimore, MD 21218, USA}%
	\thanks{$^{}$R. Kaliki and G. L\'{e}vay are with Infinite Biomedical Technologies, LLC., Baltimore, MD 21218, USA}%
	\thanks{$^{}$M. Fifer is with the Research and Exploratory Development Department, Johns Hopkins University Applied Physics Lab, Laurel, MD 20723, USA}%
	\thanks{$^{}$A. Dhawan and S. Sikdar are with the Department of Bioengineering, George Mason University, Fairfax, VA 22030, USA}%
} 

\begin{document}
	\maketitle
	  	\begin{tikzpicture}[remember picture,overlay]		
			\node[align=center,text=black] at ([yshift=-2.5em]current page text area.south) {\small This work is modified from Chapter 5 in J. L. Betthauser, “Robust Adaptive Strategies for Myographic Prosthesis Movement Decoding,”\\ \small \emph{Doctoral Dissertation}, Department of Electrical and Computer Engineering, The Johns Hopkins University, Baltimore, MD, 2020.\\ \small We provide this material to ArXiV in order to increase its visibility to the public and scientific research community.};
		\end{tikzpicture}%
	\thispagestyle{empty}
	\pagestyle{empty}
	
\begin{abstract}
	One of the most elusive goals in myographic prosthesis control is the ability to reliably decode continuous positions simultaneously across multiple degrees-of-freedom. \emph{Goal:} To demonstrate dexterous, natural, biomimetic finger and wrist control of the highly advanced robotic Modular Prosthetic Limb. \emph{Methods:} We combine sequential temporal regression models and reinforcement learning using myographic signals to predict continuous simultaneous predictions of 7 finger and wrist degrees-of-freedom for 9 non-amputee human subjects in a minimally-constrained freeform training process. \emph{Results:} We demonstrate highly dexterous 7 DoF position-based regression for prosthesis control from EMG signals, with significantly lower error rates than traditional approaches $(p<0.001)$ and nearly zero prediction response time delay $(p<0.001)$. Their performance can be continuously improved at any time using our freeform reinforcement process. \emph{Significance:} We have demonstrated the most dexterous, biomimetic, and natural prosthesis control performance ever obtained from the surface EMG signal. Our reinforcement approach allowed us to abandon standard training protocols and simply allow the subject to move in any desired way while our models adapt. \emph{Conclusions:}  This work redefines the state-of-the-art in myographic decoding in terms of the reliability, responsiveness, and movement complexity available from prosthesis control systems. The present-day emergence and convergence of advanced algorithmic methods, experiment protocols, dexterous robotic prostheses, and sensor modalities represents a unique opportunity to finally realize our ultimate goal of achieving fully restorative natural upper-limb function for amputees. 
\end{abstract} 


\section{Introduction}
\subsection{Myographic Classification, Proportional Control, and Regression}
When using machine learning methods for electromyographic (EMG) prosthesis control, there are three primary ways of decoding movements: movement classification, proportional control, and regression. EMG movement classification has a rich history, with dedicated research going back at least two decades~\cite{Hudgins93,Englehart99}. Movement classification involves selecting and predicting among a discrete set of movement types, and much of the research since has centered around which classification \emph{model} could achieve the highest accuracy from the largest set of movements. 

Proportional control~\cite{Jiang09, Cipriani11b, Simon11c} is a two-part process wherein a classification prediction is made among discrete movements and a proportion decision is made based on muscle contraction force to estimate the velocity of that predicted movement. EMG-based proportional control methods are among the most reliable and clinically viable, and have become the state-of-the-art in real-world prosthesis control~\cite{Ameri14, Fougner14, Scheme14, Amsuess15, He15, Farina17, Kanitz18}. Ultrasonic sensors have even been devised to perform extremely precise myographic proportional control~\cite{Sikdar2014, Dhawan19}.

Regression is distinct from proportional control in that there is no discrete movement decision made. Instead, a regression model seeks to estimate the continuous velocity or position of each controllable degree-of-freedom (DoF) directly from the EMG signal. Furthermore, regression control has the potential to result in movement types that are not explicitly pre-selected ahead of time. Though it has only recently been well-established that EMG regression control can outperform proportional control~\cite{Hahne18}, much research has been devoted to this EMG decoding paradigm. Regression has been used to achieve extremely reliable 2 and 3 DoF simultaneous hand/wrist prosthesis control~\cite{Pan17, Vujaklija18, Bakshi18, Ameri19}. Regression has been used to decode individual fingers independently~\cite{Tenore09, Smith08, Smith09} and simultaneously~\cite{Baldacchino18}. Recently, temporal models like long short-term memory (LSTM) recurrent networks~\cite{Hochreiter97, Graves05} have been used for regression to estimate the 3-D hand end-effector location from shoulder-recorded EMG~\cite{Xia18} and 7 independent hand and finger poses~\cite{Quivira18}.	
\begin{figure*}[t!]
	\centering
	\includegraphics[width=.99\textwidth]{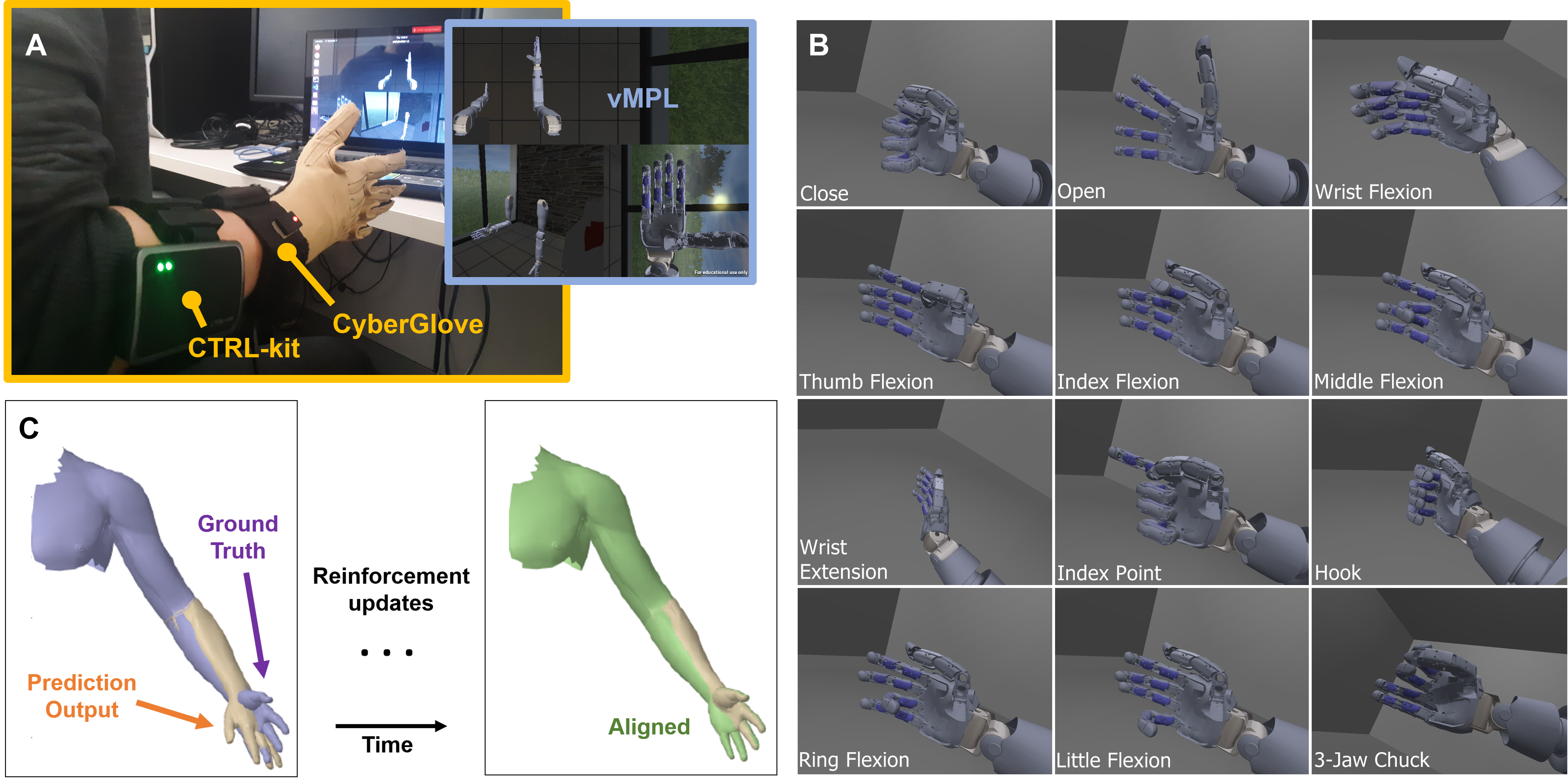}
	\caption{(A) Experiment set-up and devices. A CTRL-kit armband was placed around the circumference of the subject's forearm to record EMG, while hand and wrist positional data were recorded with a CyberGlove. The vMPL interface provided subjects with a real-time display of their hand and wrist movements on a virtual prosthesis during all of our experiments. (B) Active movement types used in the standard experiment paradigm, wherein these movements are cued to the subject, and the subjects performs each by transitioning from a rest state into the movement and back to rest. Prediction models are then trained on this data and tested on similar data. (C). Freeform reinforcement paradigm, wherein the subject performs any types of movement (not limited the those in B) s/he prefers and the prediction models are tested in real-time and periodically updated with new information. This type of training is designed to gradually improve performance by adapting to the subject in real-time, resulting in better spatial and temporal alignment/agreement between ground truth and model predictions.}
	\label{experiment_setup}
\end{figure*} 

We also wish to note another crucial aspect of using machine learning methods for EMG movement decoding: the experiment training processes. Many of the experiments cited above are highly structured and constrained, centering around the execution of a specific subset of movements that are performed as precisely as possible using significant amounts of muscle contraction force for each. This structured process is designed primarily to obtain the most distinct patterns and proportional force levels possible for each movement such that machine learning tools can readily learn to discriminate and predict the movements. While these experiment methods are effective in their own right, they require amputee users to perform motions in ways that are not representative of the way non-amputees would activate and use their intact hands and wrists.		

One of the newest ways that researchers are enhancing machine learning performance in various fields is through the use of reinforcement learning. Very recently, research groups have demonstrated the power of using reinforcement processes to achieve sophisticated dexterous robotic control~\cite{OpenAI19, Ficuciello19}, such as an automated robotic hand learning to solve a Rubik's Cube~\cite{OpenAI19}. EMG decoding research has begun to take note of the value of using adaptive prediction models~\cite{Sensinger09}, and a small number of groups have specifically integrated reinforcement training processes into their EMG prosthesis control methods~\cite{Dantas19, Kukker18, Betthauser19a}. Herein, in addition to achieving a high degree of dexterous prosthesis control, we demonstrate how reinforcement training can liberate us from the constrained and structured experiment paradigms of the past.

\subsection{Objectives}
In our previous work~\cite{Betthauser19a, Betthauser19b}, we comprehensively demonstrated that there is a tremendous benefit to using sequential models like temporal convolutional networks (TCN)~\cite{Lea16, Lea17, Kim17} for the \emph{classification} of movement intentions from EMG sequences. Specifically, these prediction models significantly improved the accuracy, stability, and transition response timing of EMG classification. Additionally, we found these sequential models particularly amenable to further improvement and human subject adaptation via a reinforcement update procedure~\cite{Betthauser19a}. At that time, our focus on classification-- as opposed to \emph{regression}-- as a vehicle for movement decoding was primarily informed by the fact that many commercial-level myoelectric prosthesis devices are controlled with pattern recognition and proportional classification strategies. In that sense, our prior work promoted research findings that could be rapidly adopted at the clinical level to yield the most immediate real-world benefits to amputees.

Our focus herein is, instead, to chart a pathway to achieving dexterous, natural, and biomimetic prosthesis control using sophisticated temporal regression prediction models combined with reinforcement learning while imposing as few constraints on the subject's movements as possible. For example, in our freeform reinforcement regression experiment, we allow the subjects to perform whatever wrist and finger movements they desire while our prediction models adapt to them. We note that the type of dexterity we are attempting to produce is designed to be implemented on advanced and highly dexterous robotic devices such as the Modular Prosthetic Limb (MPL)~\cite{Ravitz13}. Though such devices are currently cost-prohibitive and/or largely unavailable for public use, we nevertheless consider them as representative of the trajectory that future commercially-available devices will follow. Therefore, having advanced the status of current prosthesis control methods in our prior work~\cite{Betthauser18, Betthauser19a, Betthauser19b}, we now turn our attention toward experimentally demonstrating the type of natural and dexterous prosthesis control that we hope will become available in the future. 

\section{Methods}	
\subsection{Experiment Protocols}	
\subsubsection{Human Subjects}
Our experiments were conducted with protocols approved by the Johns Hopkins Medicine Institutional Review Boards. 9 non-amputee subjects (7 male, 2 female) participated in these experiments and were ages 25.3$\pm$5.5 years. All subjects except one were right-handed and performed experiments with their right arm. It should be noted that the experiments detailed herein are actively ongoing. Therefore, this complement of human subjects does not reflect the full group of non-amputee and amputee subjects who will be included in the final journal publication.

\subsubsection{Data Acquisition and Processing} 
Sixteen channels of raw EMG sampled at 2000~Hz were obtained from a CTRL-kit armband (CTRL-Labs, New York, NY) placed around the circumference of the forearm (Fig.~\ref{experiment_setup}A). Five standard EMG time-domain (TD5) features were extracted in real-time from the raw EMG signals: mean absolute value, waveform length, variance, slope sign change, and zero crossings~\cite{Hudgins93}. The optimal window sizes used were  different for each prediction model (Table~\ref{model_params_table}), a point we will discuss later. Hand and wrist positional data were recorded with a CyberGlove II (CyberGlove Systems LLC, San Jose, CA). We built a user interface to control the virtual Modular Prosthetic Limb (vMPL) developed by the Johns Hopkins University Applied Physics Laboratory~\cite{Ravitz13}. The vMPL interface provided subjects with a real-time display of their hand and wrist movements during all of our experiments by approximating the dynamic properties of the physical robotic Modular Prosthetic Limb (MPL). Subjects wore both sensors on either the left or right arm depending on preference.

\subsubsection{Ground Truth Encoding}
As a calibration procedure at the beginning of each experiment, subjects explored for 15~s the full range of finger and wrist motions in each of 7 degrees-of-freedom (DoF): flexion/extension of 5 fingers, flexion/extension of the wrist, and adduction/abduction of the wrist. For each DOF $i$, we recorded the minimum and maximum CyberGlove values, $\rho^{min}_i$ and $\rho^{max}_i$, obtained during this calibration. We then used these ranges to create two scale mappings: one to produce angular output to the vMPL interface, $\rho\in[\rho^{min}_i,\rho^{max}_i]\rightarrow\theta_i\in[\theta^{min}_i,\theta^{max}_i]$, and one to produce normalized values for model training, $\rho\in[\rho^{min}_i,\rho^{max}_i]\rightarrow\phi_i\in[0,1]$. Thereafter, at every 25~ms time-step during our experiments, we obtained EMG features, 7 DoF angular position data, and 7 DoF normalized angular position data.

\subsubsection{Standard Experiment: Validation of Basic Regression}
We thought it prudent to first carry out what we will call a \emph{standard} type of EMG decoding experiment, often used for proportional classification tasks in research and clinical settings, to test the value of our methods for 7 DoF position-based regression in this context. The central feature of the \emph{standard} paradigm is this: that specific movement cues are presented to the subject and the subject performs repetitions of those-- and only those-- movements, typically alternating from a rested state into the specified active movement state and back to rest, and the movements are performed at a relatively high fraction of a subject's maximum voluntary contraction force. Variations on this paradigm can be included, such as allowing for adjustments in the velocity and/or muscle contraction effort used to perform the movement types. In terms of our ultimate goal of achieving natural biomimetic prosthesis control, this standard experiment paradigm is problematic in a number of ways, but we have already touched on three. First, the cued movements are a pre-selected limited set. Models are trained on this limited set, and no other movement types will occur at run-time. Second, as mention above, transitions are limited to those to and from a rest state, implying that prediction models will learn little if any information about how to transition from one active state directly to another. Third, movements are performed using relatively strong muscle contraction forces that are highly fatiguing for the subject. As a result, standard experiment paradigms often have large periods of rest built into their protocols due to muscle fatigue.  Biomimetic human movement is in reality a continuum of motions rather than discrete state transitions and, consequently, the standard training paradigm applied to prosthesis control cannot be expected to result in natural biomimetic movement.

\begin{figure*}[t!]
	\centering
	\includegraphics[width=.9\textwidth]{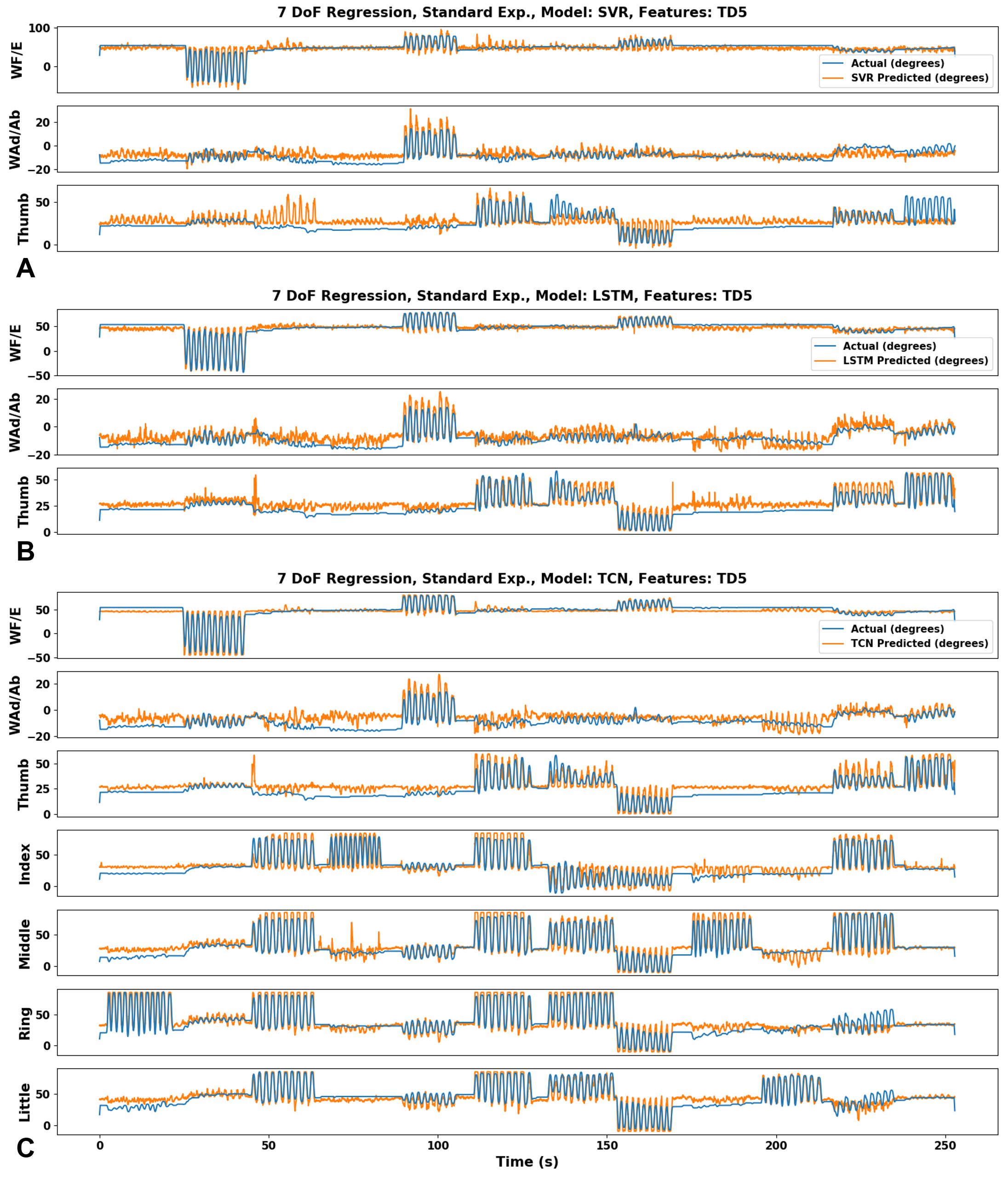}
	\caption{Standard experiment offline regression output for a single subject when predicting using the following models: (A) SVR (B) LSTM, and (C) TCN. TCN is our preferred model so we show its simultaneous output for all 7 DoFs, whereas we only show the first 3 DoFs for SVR and LSTM. These prediction output profiles are provided to give a sense of the behavioral characteristics each model for granular comparison. For example, we note that the total error of LSTM was slightly lower than TCN, but LSTM prediction outputs also tended to be visibly noisier than TCN.}
	\label{standard_output}
\end{figure*}

All nine non-amputee subjects participated in this experiment. Each subject participated in a 15~min introductory session wherein the experiment and 12 active movement types were explained (Fig.~\ref{experiment_setup}B). The subjects then practiced performing these movements. Once s/he was comfortable performing them, this first experiment was initiated. For 15~min, each subject preformed 3 trials of these movements when cued. For a given trial, each active movement was cued exactly once for in a random order, alternating between 20~s cues for an active movement and 5~s cues for rest. During the 20~s when an active cue was presented, the subject was allowed to perform that movement multiple times by transitioning from rest into the movement and back to rest. The subject was instructed at the outset that s/he could vary the velocity, but to only apply a normal contraction effort level necessary to achieve the cued movements at their preferred velocities. Our preference for minimal muscular contraction effort in our experiments are distinct from many classification and proportional prosthesis activation methods in which much higher force contraction levels are used. During the 5~s when a rest cue was presented, the subject rested their hand and wrist. 

To obtain our results (Fig.~\ref{standard_output},~\ref{angular_error}A, and~\ref{temporal_correlation}A), we used the first 2 trials of each subject's data to train prediction models, and those models were tested on the last trial of data. Model parameters in Table~\ref{model_params_table} were determined from parameter sweeps using preliminary data obtained prior to these experiments.  

\subsubsection{Freeform Dexterous Movement Reinforcement Experiment}	
As in our prior investigation with reinforcement adaptation to improve movement \emph{classification}~\cite{Betthauser19a}, with the experiment below we seek to improve position-based \emph{regression} error over time by adapting to each subject's naturally preferred and evolving behavior through unguided human-machine interaction. In this way, we are shifting away from standard or traditional guided training protocols toward the idea of having no specific protocol whatsoever. Instead, what we would like to promote is a ``learn by doing'' approach enabled through reinforcement learning wherein the subject simply executes motions that s/he naturally prefers and the prediction models adapt by incorporating new information and previously unseen movements. The subject can resume this human-machine interaction at any time to continue refining and optimizing the prediction model.	

Nine non-amputee subjects participated in this experiment, though one subject was excluded because his data were not properly saved to file at run-time. The experiment was sectioned into a 1~min training trial to initialize our prediction models and fifteen consecutive 30~s test and update trials for a total experiment time of around 9~min. For the total duration of the experiment, subjects explored any continuous movements and combinations along the specified 7 DoFs in any way they desired. Subjects were only instructed to perform movements in ways they naturally would, in freeform open-ended fashion, with the natural contraction effort levels necessary to achieve their preferred movements and speed. At the end of first 1~min trial, the data obtained was used to initialize prediction models. For each trial thereafter, the subject was then able to observe real-time output predictions from the TCN sequential model in the vMPL interface, whereas LSTM model predictions were simply conducted in the background and preserved for later analysis. Specifically, after training the models with data from the first trial, this data was discarded (though it was saved to a file for later analysis). The models were then tested for the next 30~s trial. The models were then further trained with a reinforcement update using the data from the next trial, these data were also discarded, and this process repeated for a total of fifteen reinforcement trials. 	

\begin{figure*}[t!]
	\centering
	\includegraphics[width=.9\textwidth]{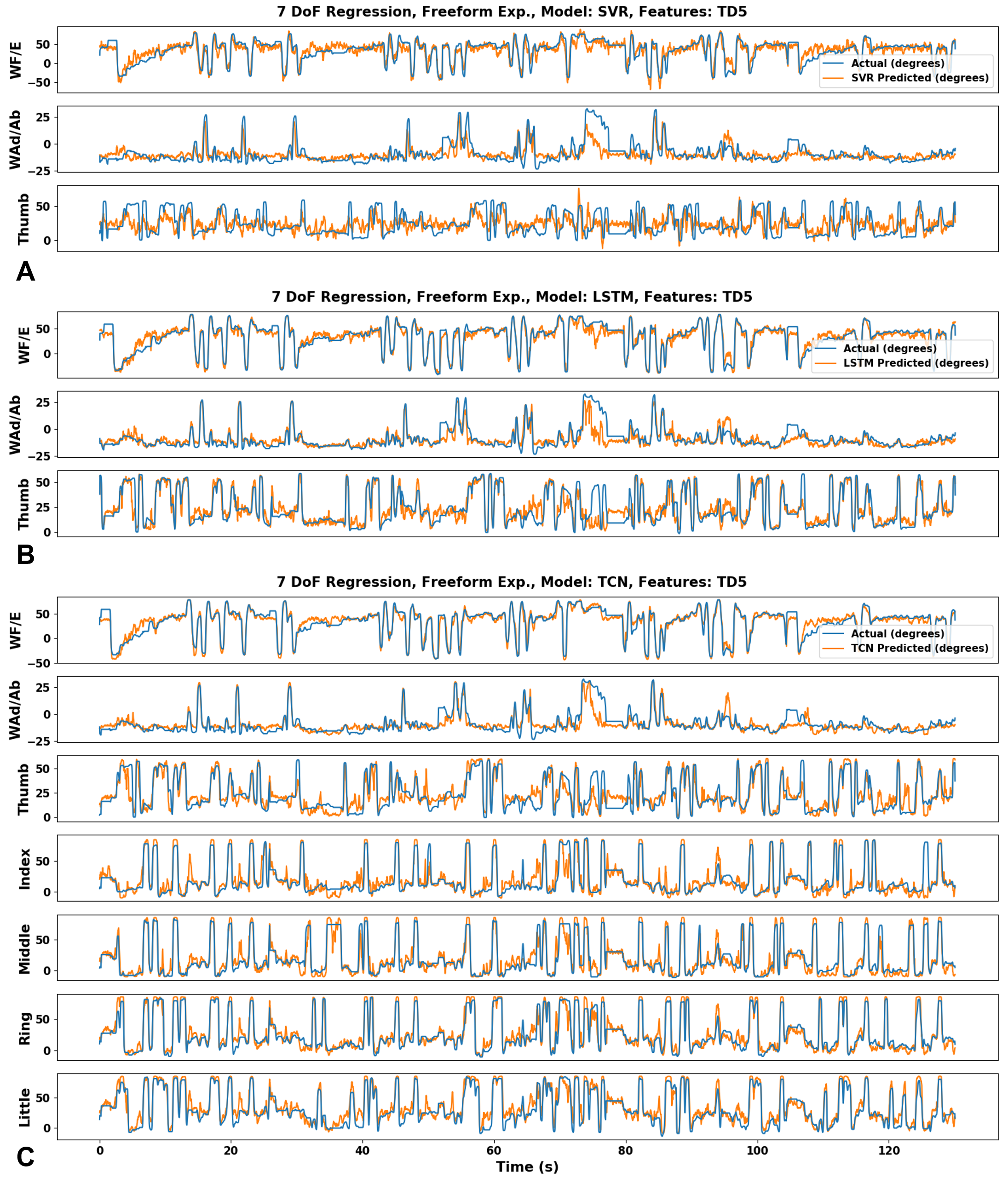}
	\caption{Freeform experiment offline regression output for a single subject when predicting using the following models: (A) SVR (B) LSTM, and (C) TCN. TCN is our preferred model so we show its simultaneous output for all 7 DoFs, whereas we only show the first 3 DoFs for SVR and LSTM. These prediction output profiles are provided to give a sense of the behavioral characteristics each model for granular comparison. For example, we note that the total error of LSTM was slightly lower than TCN, but LSTM prediction outputs also tended to be visibly noisier than TCN.}
	\label{freeform_output}
\end{figure*}

We performed two distinct types of analyses on the data obtained in this experiment. (i) The first analysis was a \emph{freeform} offline assessment using each subject's entire 9~min of data wherein we used the first 60\% of data to train prediction models, and those models were tested on the last 40\% of data (Fig.~\ref{freeform_output},~\ref{angular_error}B, and~\ref{temporal_correlation}B). (ii) The second analysis was an online \emph{reinforcement} assessment of model regression performance during the experiment's run-time (Fig.~\ref{reinforce_results}).

\subsubsection{Prediction Models, Performance Assessment, and Optimal Parameters}
\begin{figure*}[t!]
	\centering
	\includegraphics[width=.99\textwidth]{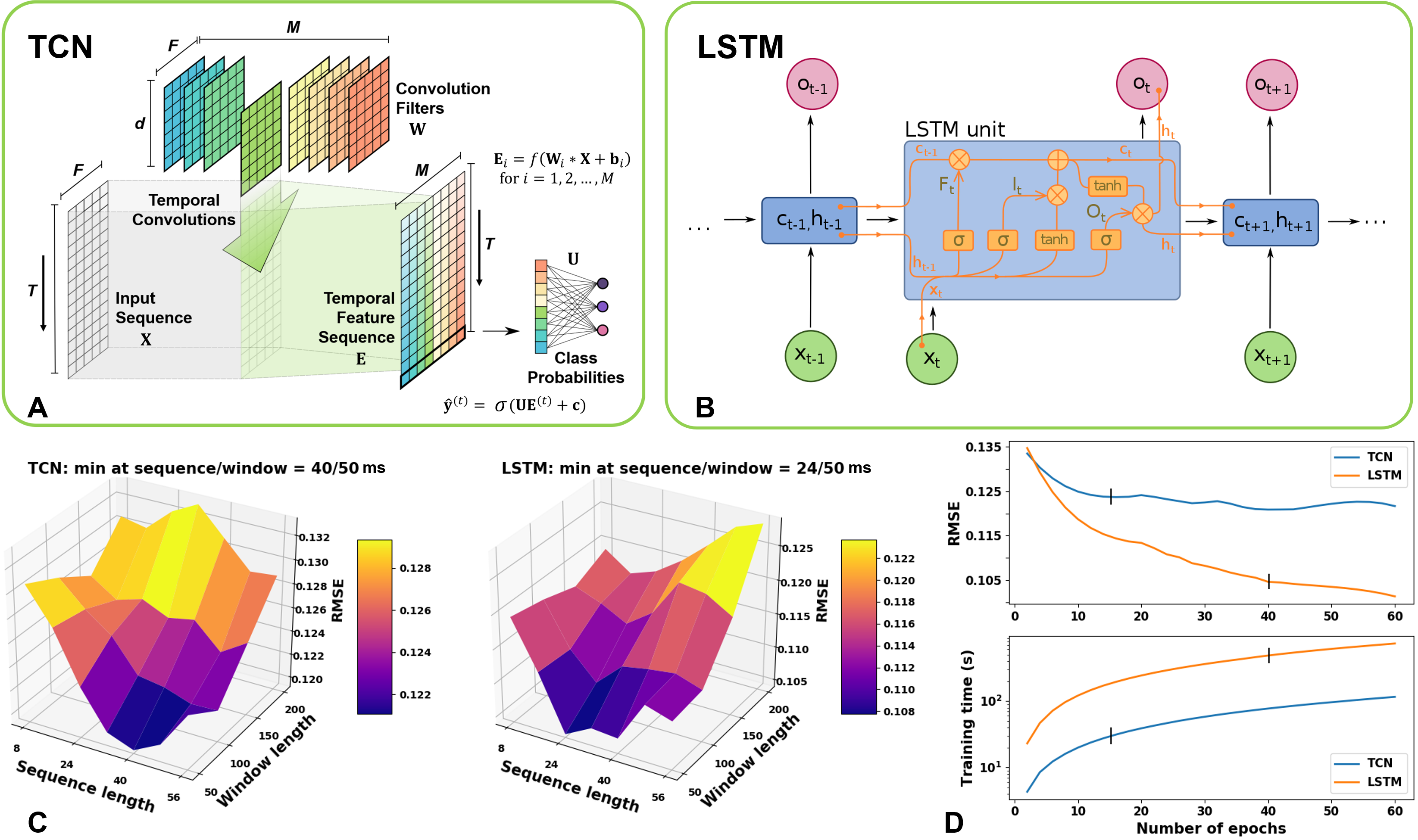}
	\caption{EMG features can be fed as fixed-length sequences into sequential prediction models such as (A) TCN (\copyright 2019 IEEE) and (B) LSTM (\copyright Wikimedia Commons). (C) Parameter-sweeps to determine optimal window sizes and sequence lengths. The inclusion of sequence data lowered error rates, and sequential models required much shorter feature windows to achieve superior performance than the typically-used 200~ms window for decoding EMG with frame-wise models. (D) RMSE and training time as a function of training epochs. We used 15 training epochs for TCN, and 40 training epochs for LSTM. Final parameters used in our experiments are shown in Table~\ref{model_params_table}. }
	\label{model_params}
\end{figure*}
To obtain our results, we used the following sequential and frame-wise models: 
\begin{table}[h!]\centering
	\begin{tabular}{@{}ll@{}}
		\midrule
		TCN: & Temporal convolutional network~\cite{Betthauser19b,Lea16}, $M$=128, $d$=25 \\
		LSTM: & Long short-term memory network~\cite{Hochreiter97}, 64 nodes\\
		SVR: & Support vector regression~\cite{Ameri14}, Gaussian radial basis, $\gamma=0.01$, $C=1$\\ \hline
	\end{tabular}
\end{table}	

\noindent The final output layer for TCN and LSTM sequential models (Fig.~\ref{model_params}A and B) were sigmoid activations, and these models were trained using mean-squared error as their loss function. SVR was chosen as robust frame-wise prediction model which has demonstrated generally good performance in multi-DoF EMG regression tasks. We tested all sequential models using TD5 feature windows and sequences. All models were constrained to be strictly causal such that predictions were based on EMG information preceding each prediction. Model parameters for window sizes, sequence lengths, and training epochs (Table~\ref{model_params_table}) were determined from parameter sweeps using preliminary data obtained prior to the experiments described herein.

Computations were performed with a GeForce 930M Laptop GPU (NVIDIA, Santa Clara, CA) using standard Python 3.7.4 libraries and the following open-source packages: \emph{Keras}~\cite{Keras18} and \emph{Temporal Convolutional Networks}~\cite{TCN18}. We computed the following metrics for model evaluation: root-mean-squared error (RMSE) and coefficient of determination ($r^2$) were used to assess each model's fidelty to ground truth, and model response time delay was computed by determining the temporal shift required for each model to achieve maximum correlation with ground truth. Kruskall-Wallis one-way variance analysis~\cite{Kruskall52} was used to compute all statistical $p$-values. Shaded regions in figures represent standard error of the mean. 

\begin{table}[t!]\centering
	\caption{Model Parameters Used for Experiments}
	\begin{tabular}{@{}ccccc@{}}	
		\emph{Model}$^{\dagger}$ &\emph{Features} & \emph{Window}& \emph{Sequence} & \emph{Epochs}\\
		\midrule		
		{TCN} & TD5 & 50 ms & 40 & 15\\
		{LSTM} & TD5 & 50 ms & 24 & 40\\ 
		{SVR} & TD5 & 200 ms & 1 & --\\ \midrule
		\multicolumn{5}{l}{$^{\dagger}$Prediction time-step: $\Delta t$ = 25~ms. Batch size: 24.}
	\end{tabular}
	\label{model_params_table}
\end{table} 

\section{Results}	
\subsection{Standard Experiment Regression}
\begin{figure*}[t!]
	\centering
	\includegraphics[width=.99\textwidth]{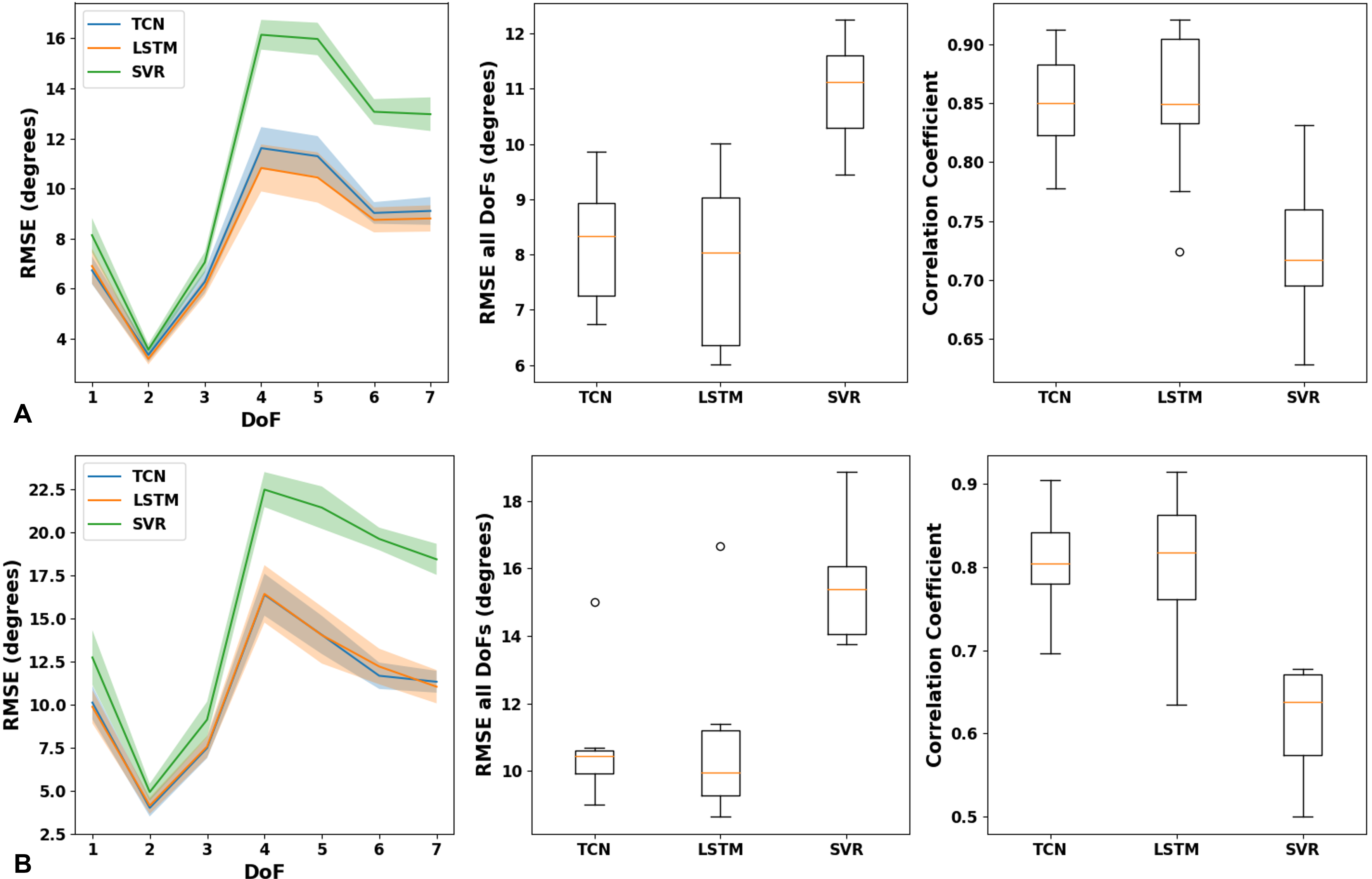}
	\caption{The per-DoF angular RMSE, total angular RMSE, and $r^2$ coefficient of determination based on regression performance across subjects during (A) the standard experiment and (B) the freeform experiment. In both experiments, the average RMSE values were better for wrist (DoFs 1 and 2) and thumb (DoF 3) motions than they were for individual fingers (DoFs 4 to 7). In both experiments, the sequential models TCN and LSTM performed similarly, and they significantly outperformed the SVR frame-wise model.}
	\label{angular_error}
\end{figure*} 
Standard experiment results are shown in Fig.~\ref{standard_output}, Fig.~\ref{angular_error}A, and Fig.~\ref{temporal_correlation}A. We note that these experiments are ongoing, and the final results we obtain may differ from those shown herein. Across 9 subjects, the average angular RMSE was 8.4$^\circ$ for TCN, 8.0$^\circ$ for LSTM, and 11.3$^\circ$ for SVR. Generally speaking, the average RMSE values were worse for individual fingers (DoFs 4 to 7) than for wrist (DoFs 1 and 2) and thumb (DoF 3) motions. The average $r^2$ correlation between ground truth and model predictions was 0.85 for TCN and LSTM, and 0.72 for SVR. For both metrics, the distribution of SVR (frame-wise) performance across subjects was significantly different ($p<0.001$) from sequential models TCN and LSTM (Fig.~\ref{angular_error}A). Furthermore, the average prediction response delays for both TCN and LSTM were less than the 25~ms prediction time-step, and were significantly faster ($p<0.01$) than the response delay of SVR (Fig.~\ref{temporal_correlation}A).

\subsection{Freeform Dexterous Movement Regression}
\begin{figure*}[t!]
	\centering
	\includegraphics[width=.99\textwidth]{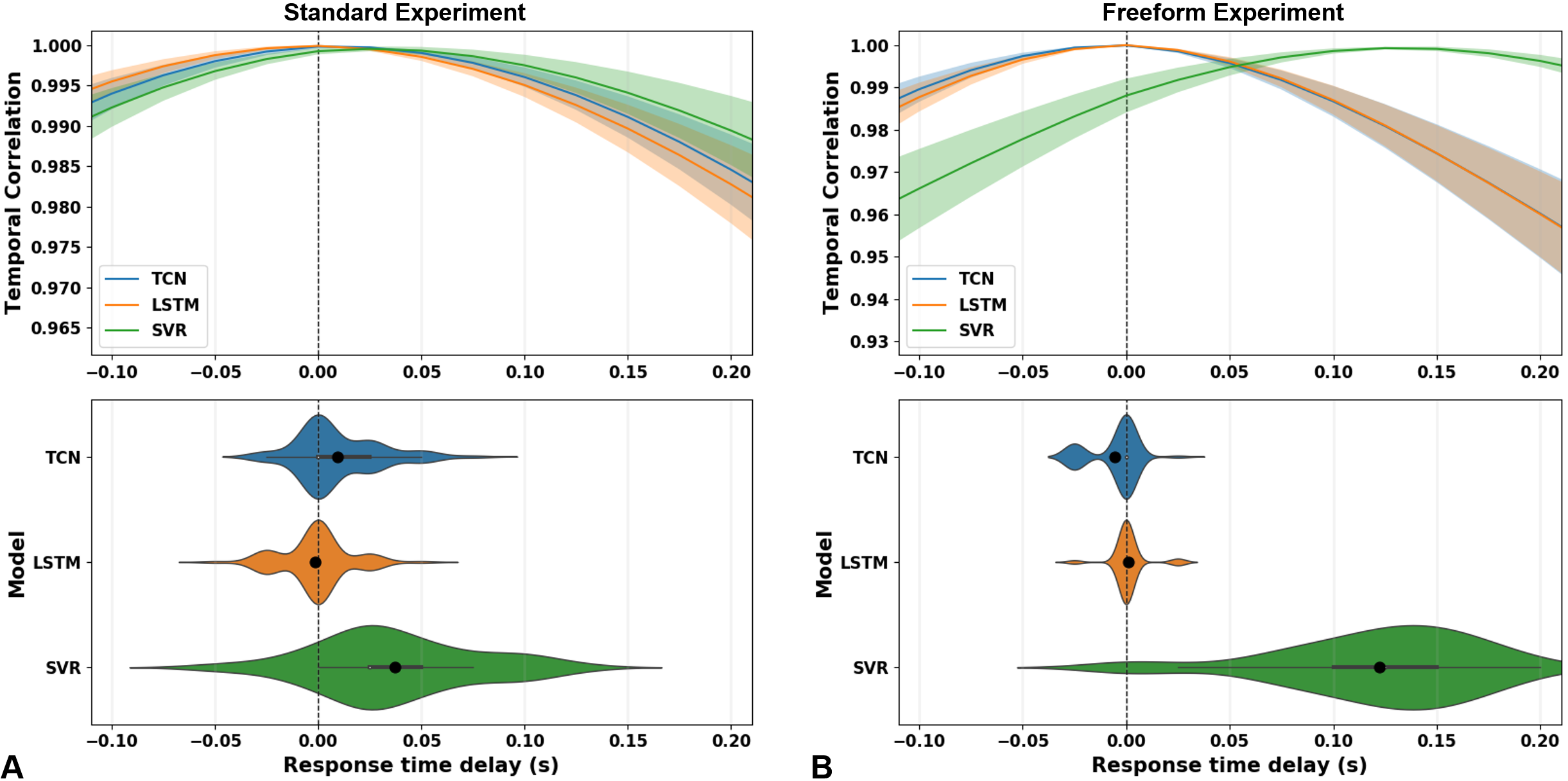}
	\caption{The normalized temporal correlation and prediction response delays for each prediction model during (A) the standard experiment and (B) the freeform experiment. In both experiments, the average maximum temporal correlation for the TCN and LSTM sequential models was within a single prediction time-step of being in phase with ground truth. In other words, there was almost no response time delay between the actual movements each subject attempted and the continuous movement predictions of the sequential models. Across subjects and DoFs, these response delay distributions were consistently tightly clustered at or around zero delay, especially as a result of the freeform training process. In contrast, the frame-wise SVR model made more delayed predictions, especially during the unconstrained movements that occurred during the freeform experiment. Given that there is a human in the loop in real-time prosthesis control, longer delays can result in negative control feedback effects such as target position over-shooting and consequent compensatory movements from the user.}
	\label{temporal_correlation}
\end{figure*}

Freeform experiment results are shown in Fig.~\ref{freeform_output}, Fig.~\ref{angular_error}B, and Fig.~\ref{temporal_correlation}B. We note that these experiments are ongoing, and the final results we obtain may differ from those shown herein. Across 8 subjects, the average angular RMSE was 10.4$^\circ$ for TCN, 9.9$^\circ$ for LSTM, and 15.4$^\circ$ for SVR. As in the previous experiment, the average RMSE values were worse for individual fingers (DoFs 4 to 7) than for wrist (DoFs 1 and 2) and thumb (DoF 3) motions. The average $r^2$ correlation between ground truth and model predictions was 0.80 for TCN, 0.82 for LSTM, and 0.64 for SVR. For both metrics, the distribution of SVR (frame-wise) performance across subjects was significantly different ($p<0.001$) from sequential models TCN and LSTM (Fig.~\ref{angular_error}B). Furthermore, the average prediction response delays for both TCN and LSTM were less than the 25~ms prediction time-step, with nearly zero-delay, and were significantly faster ($p<0.001$) than the 120~ms average response delay of SVR (Fig.~\ref{temporal_correlation}B). 

\subsection{Reinforcement Learning during Freeform Exploration}
\begin{figure*}[t!]
	\centering
	\includegraphics[width=.99\textwidth]{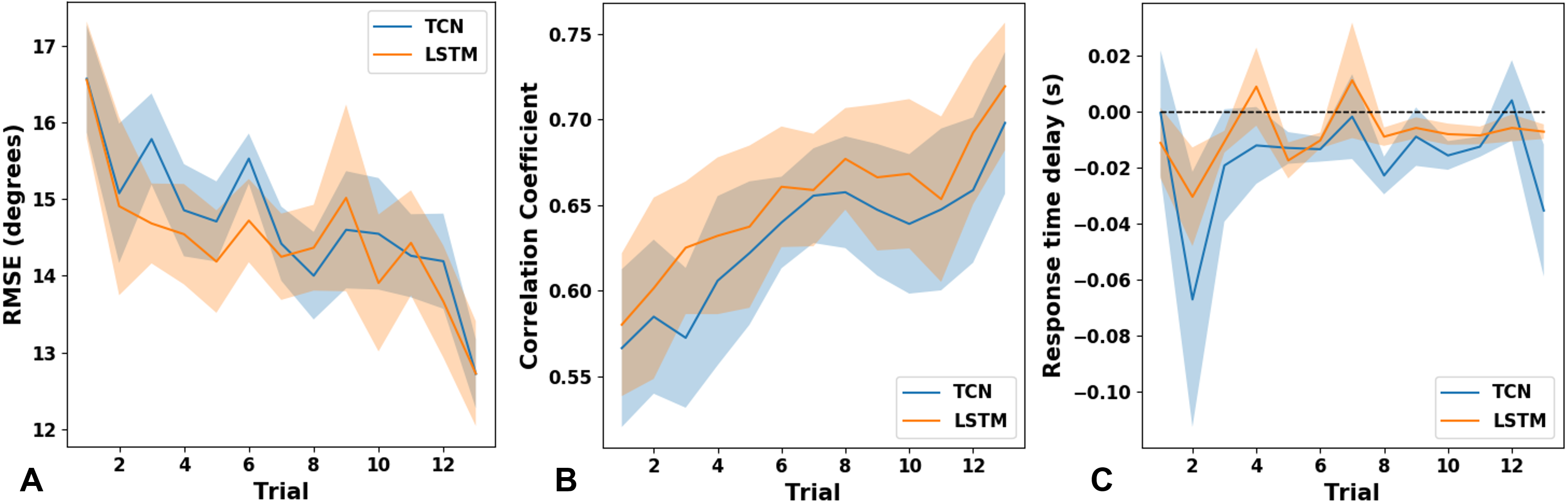}
	\caption{Sequential model performance during the online reinforcement experiment in terms of (A) angular RMSE, (B) $r^2$ coefficient of determination, and (C) prediction response time delay. TCN and LSTM performed quite similarly when compared to each other. By the end of the experiment, both prediction models had improved significantly from their behavior at the beginning of the experiment. Sequential model predictions are only slightly out of temporal phase with respect real-time ground truth whereas, in the past, typical response time delays have tended to exceed 100~ms.}
	\label{reinforce_results}
\end{figure*}

Online reinforcement experiment results are shown in Fig.~\ref{reinforce_results}. We note that these experiments are ongoing, and the final results we obtain may differ from those shown herein. The TCN and LSTM sequential models are naturally amenable to a reinforcement training process~\cite{Betthauser19a}, allowing for  update training on the fly and, subsequently, the permanent discarding of past data. Thus, the sequential models were the only ones used for this experiment. We were not seeking to do a direct model comparison analysis with this experiment, but merely to highlight the utility and benefits of using reinforcement learning to train sequential regression models. Indeed, Fig.~\ref{reinforce_results} demonstrates that TCN and LSTM performed quite similarly when compared to each other. By the end of the experiment, both prediction models had improved significantly ($p<0.001$) from their behavior at the beginning of the experiment. Angular RMSE had decreased by about 4$^\circ$ (Fig.~\ref{reinforce_results}A), and $r^2$ correlation improved by about 0.13 (Fig.~\ref{reinforce_results}B). While we are unable at this time to draw a statistical inference about the evolution of model response time delay, we can observe that the model predictions are only slightly out of phase with the real-time ground truth (Fig.~\ref{reinforce_results}B). Typical response time delays in EMG prosthesis control models tend to exceed 100~ms~\cite{Smith11}.

\section{Discussion}
As previously mentioned, the experiments described herein are ongoing so we want to be cautious at this time about stating any hard conclusions. However, we can make some observations about the results we have obtained thus far. Foremost among them is the quality of sequential model regression performance, even in the minimally-constrained freeform experiment context, and even with minimal muscular force or effort expenditure by the subjects. Typically, experiments are highly structured and constrained, centering around the execution of a specific subset of movements (Fig.~\ref{experiment_setup}B) that are performed as precisely as possible using a large fraction of the subject's maximum voluntary contraction (MVC) force for each. This structured process is designed primarily to obtain the most distinct patterns and proportional force levels possible for each movement such that machine learning tools can readily learn to discriminate and predict the movements. Given these constraints, there have been numerous successful techniques at the research and commercial levels to achieve 2 or 3 DoFs of simultaneous and proportional prosthesis control. On the other hand, these techniques require amputee users to perform motions in ways that are not representative of the way non-amputees would activate and use their intact hands and wrists. In a biomimetic sense, these movements are performed in unnatural fashion and require muscle contraction forces that are highly fatiguing for the subject. Indeed, many EMG prosthesis control experiments often have large periods of rest built into their protocols precisely because muscle fatigue is such a major factor. 

To achieve the ultimate goal of dexterous and biomimetic prosthesis control, we must liberate ourselves from such limitations. Prior research has demonstrated RMSE of around 5$^\circ$-6$^\circ$ for 3 DoF control tasks using the experimental constraints described above. In our standard constrained experiment, our models achieved roughly 8$^\circ$ RMSE for 7 simultaneous DoF positional regression, even while allowing and encouraging subjects to use minimal muscle contraction forces (Fig.~\ref{angular_error}). In our minimally-constrained freeform experiment, allowing the subjects to do any 7 Dof finger and wrist movements that they desired at even minimal muscle contraction forces, our models achieved about 10$^\circ$ RMSE for regression.  We've already shown how sequential models achieve nearly real-time EMG classification predictions, whereas the prediction delay of frame-wise models tend to be quite high, being governed primarily by feature window size and choice of post-processing methods~\cite{Betthauser19a}. The regression results herein are consistent with those findings, with sequential models achieving nearly real-time temporal correlation to ground truth (Fig.~\ref{temporal_correlation}). In other words, there was almost no response time delay between the actual movements each subject attempted and the continuous movement predictions of the sequential models. Across subjects and DoFs, these response delay distributions were consistently tightly clustered at or around zero delay, especially during the freeform training experiment. This result is notable because the sequential models learned the best temporal synchronization with ground truth via the freeform experiment process rather than an arguably simpler standard experiment. In contrast, the frame-wise SVR model made more delayed predictions, especially during the unconstrained movements that occurred during the freeform experiment. Given that there is a human in the loop in real-time prosthesis control, longer delays can result in negative control feedback effects such as target position over-shooting and consequent compensatory movements from the user. 

Our freeform and reinforcement experimental approaches are fundamentally different from previous experiments and, as such, we will fore-go making any additional comparative analysis among models or prior research (Fig.~\ref{reinforce_results}). There will always be another newer model to compare, but we are instead demonstrating and defining an entirely new way of thinking about the problem of prosthesis control. Using our algorithmic and ``learn by doing'' experimental methods, we have demonstrated a degree of dexterous, biomimetic, and natural prosthesis control to a degree that has never before been realized (Fig.~\ref{standard_output} and \ref{freeform_output}). The supplemental videos we will add to this work will demonstrate and emphasize even more effectively than our figures just how human-natural these prediction outputs are.

\subsection{Sonomyography Discussion}
\begin{figure*}[t!]
	\centering
	\includegraphics[width=.9\textwidth]{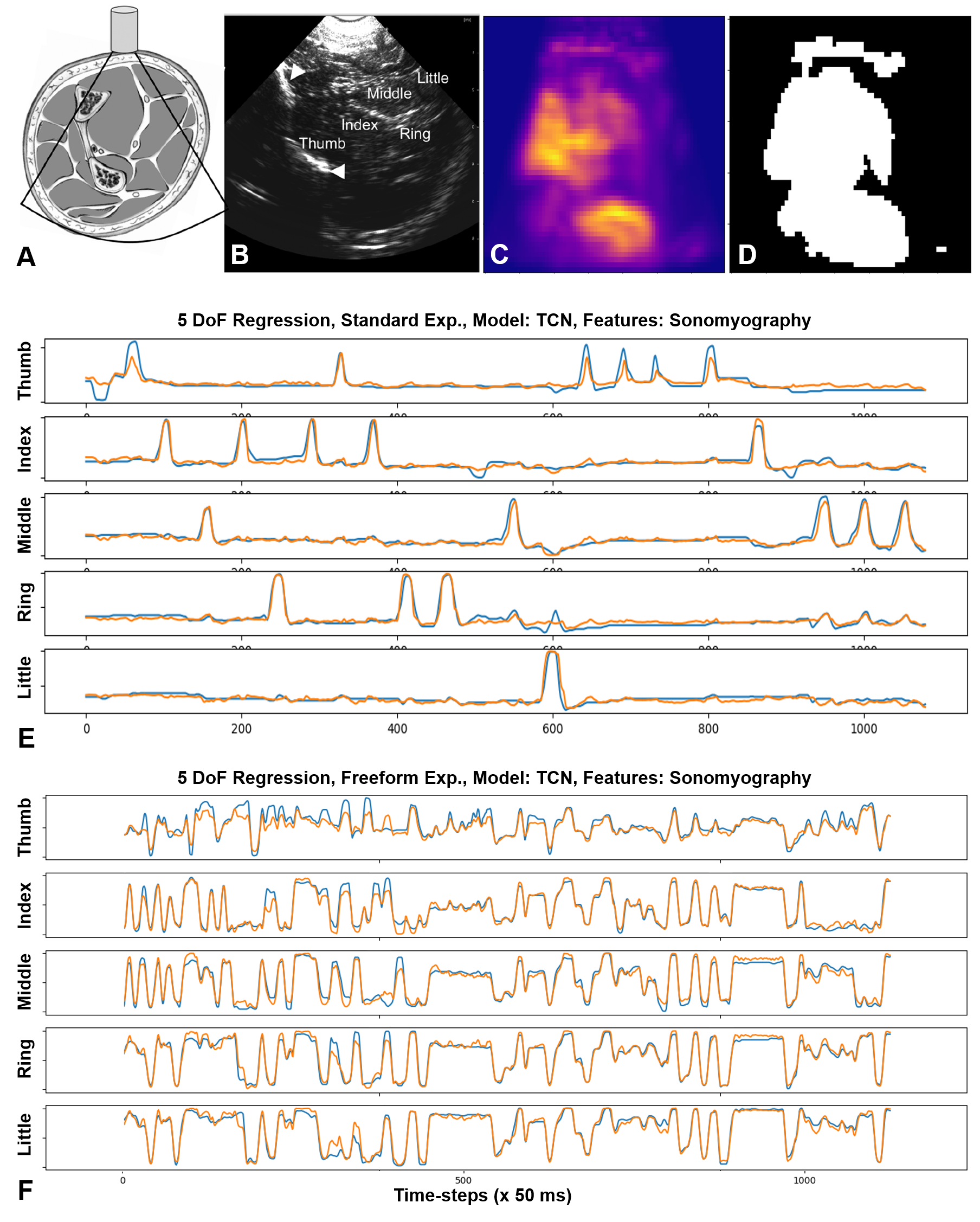}
	\caption{(A) Conceptual illustrated cross-section of the human forearm and a low-power sonomyographic sensor for recording muscle movements. \copyright 2014 IEEE (B) Actual image of the human forearm as recorded from the sonomyographic sensor. \copyright 2014 IEEE (C) Since sonomyography is a relatively high-dimensional signal, we elected to down-sample the images and apply Gaussian smoothing before inputting them into our sequential models. We also compute the per-pixel variance (shown) across the training set (D) and use the regions of highest variance to create a binary mask to further reduce the input size. For this subject, the input dimensionality was reduced by this process by a factor of roughly 12. We used the TCN model to perform regression from sonomyographic inputs to control 5 finger DoFs (E) independently in similar fashion to our standard experiment and (F) in unconstrained freeform fashion. The sonomyographic sensor results in very smooth and accurate prediction streams and we expect it to eventually replace EMG as the dominant myographic modality.}
	\label{sono_results}
\end{figure*}

The experiments that remain for this work are centered around real-time outcomes measures-based assessments of our methods for amputees such as target achievement control testing~\cite{Simon_2011} and implementation on the physical MPL arm. We are also closely coordinating our work with a group at George Mason University who have developed a next-generation ultrasound-based myographic sensor (sonomyography) for prosthesis control~\cite{Sikdar2014}\cite{Dhawan19}. We have applied our methods with their sonomyographic sensor and obtained extremely promising preliminary results (Fig.~\ref{sono_results}). In fact, we believe it will one day become the dominant myographic sensor modality. We will be appending a comparative analysis for electromyography and sonomyography using these methods in order to emphasize the multi-modal flexibility of our methods and to further illuminate what we believe to be the future of prosthesis control technology.

\section{Conclusions}
We have now demonstrated highly dexterous multi-DoF position-based regression for prosthesis control from EMG signals. Our reinforcement approach allowed us to abandon standard training protocols and simply allow the subject to move in any desired way while our models adapted. To our knowledge, the methods described herein elicit the most dexterous, biomimetic, and natural prosthesis control performance ever obtained from the surface EMG signal. As such, this work redefines the state-of-the-art in myographic decoding in terms of the reliability, responsiveness, and movement complexity available from prosthesis control systems. The present-day emergence and convergence of advanced algorithmic methods, experiment protocols, dexterous robotic prostheses, and sensor modalities represents a watershed moment for our field. We are on the cusp, and now have a unique opportunity to finally realize our ultimate goal of achieving fully restorative natural upper-limb function for amputees. 

\section*{Acknowledgment} 
We thank the Johns Hopkins Applied Physics Laboratory (JHU/APL) for making available the vMPL, developed under their Revolutionizing Prosthetics program and based upon work supported by the Defense Advanced Research Projects Agency (DARPA) under Contract No. N66001-10-C-4056. Any opinions, findings and conclusions or recommendations expressed in this material are those of the author(s) and do not necessarily reflect the views of DARPA and JHU/APL. We thank the human subjects who participated in this study and our colleagues Dr. Brock Wester, Robert Armiger, Dr. Colin Lea, Tae Soo Kim, and Dr. Austin Reiter. 



\bibliographystyle{ieeetr}

\end{document}